\documentclass{article}

\usepackage{amsmath}
\usepackage{graphicx} 
\usepackage{subfigure}
\usepackage{algorithm}
\usepackage{algorithmic}
\usepackage[final,nonatbib]{nips_2017}

\usepackage{tikz}
\usepackage[utf8]{inputenc} % allow utf-8 input
\usepackage[T1]{fontenc}    % use 8-bit T1 fonts
\usepackage{hyperref}       % hyperlinks
\usepackage{url}            % simple URL typesetting
\usepackage{booktabs}       % professional-quality tables
\usepackage{amsfonts}       % blackboard math symbols
\usepackage{nicefrac}       % compact symbols for 1/2, etc.
\usepackage{microtype}      % microtypography

\title{Generative Adversarial Active Learning}

\author{
  Jia-Jie Zhu\\
 	Max Planck Institute for Intelligent Systems\\
  T\"ubingen, Germany\\
  \texttt{jia-jie.zhu@tuebingen.mpg.de} \\
  \And
  Jose Bento\\
  Department of Computer Science\\
  Boston College\\
  Chestnut Hill, Massachusetts, USA\\
  \texttt{jose.bento@bc.edu} \\
}

\begin{document}

\maketitle

\begin{abstract}
We propose a new active learning by query synthesis approach using Generative Adversarial Networks (GAN). 
Different from regular active learning, the resulting algorithm adaptively synthesizes training instances for querying to increase learning speed. 
We generate queries according to the uncertainty principle, but our idea can work with other active learning principles.
We report results from various numerical experiments to demonstrate the effectiveness the proposed approach.
In some settings, the proposed algorithm outperforms traditional pool-based approaches.
To the best our knowledge, this is the first active learning work using GAN.
\end{abstract}

\section{Introduction}
\label{intro}
One of the most exciting machine learning breakthroughs in recent years is the generative adversarial networks (GAN) \cite{goodfellow2014generative}. It trains a generative model by finding the Nash Equilibrium of a two-player adversarial game.
Its ability to generate samples in complex domains enables new possibilities for active learners to synthesize training samples on demand, rather than relying on choosing instances to query from a given pool.

\begin{figure}[ht]
	\centering
	\subfigure[Pool-based]{%
		\includegraphics[width=0.2\columnwidth]{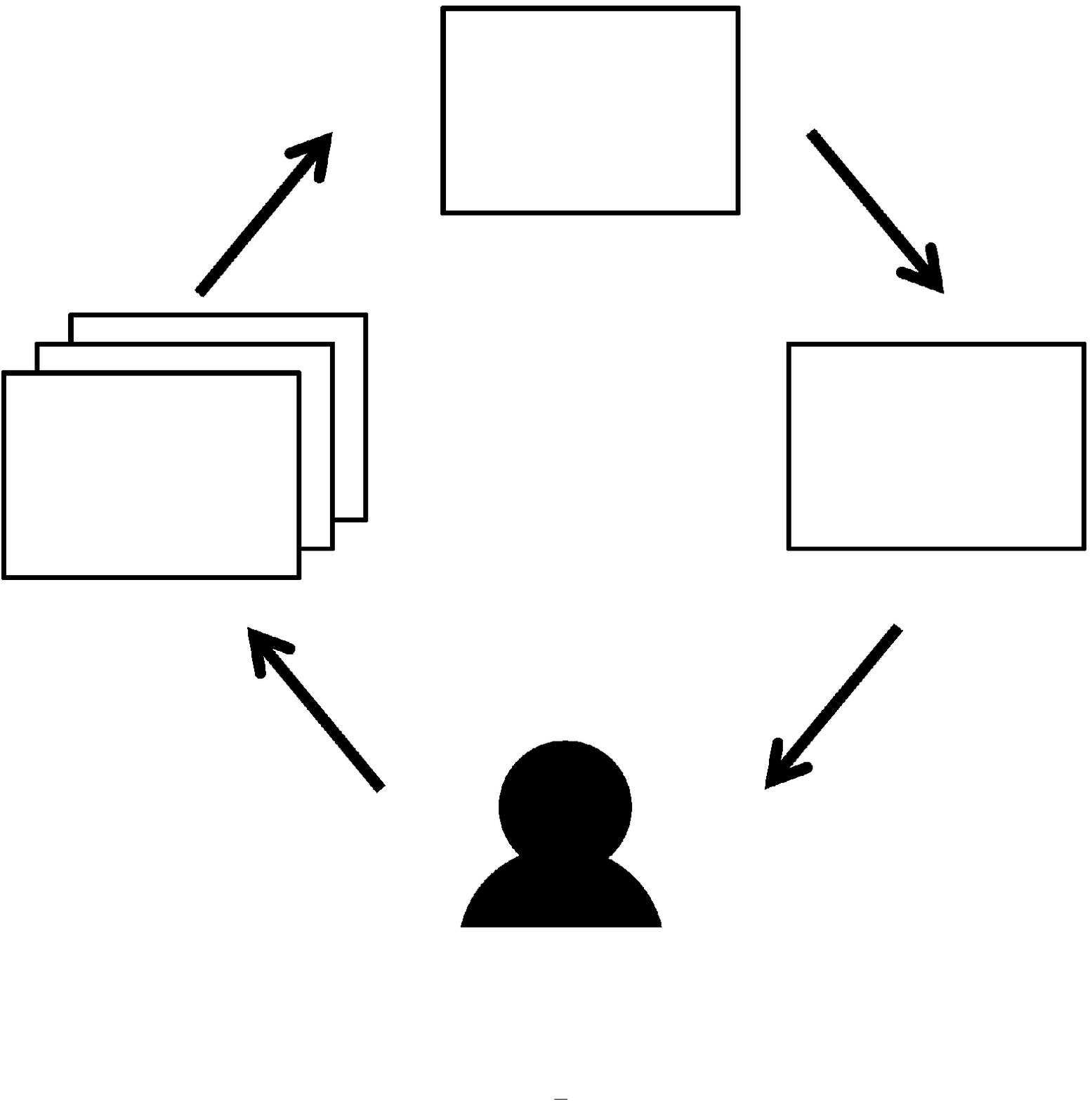}
		\put(-50,83){{\footnotesize Learner}}
		\put(-19,45){{\footnotesize Pool}}
		\put(-116,65){{\footnotesize Training}}
		\put(-15,20){{\footnotesize $x,?$}}
		\put(-75,20){{\footnotesize $x,y$}}
	}\;\;\;\;\;\;\;\;\;\;\;\;\;\;\;\;\;\;\;\;\;\;
	\subfigure[GAAL]{%
		\includegraphics[width=0.18\columnwidth]{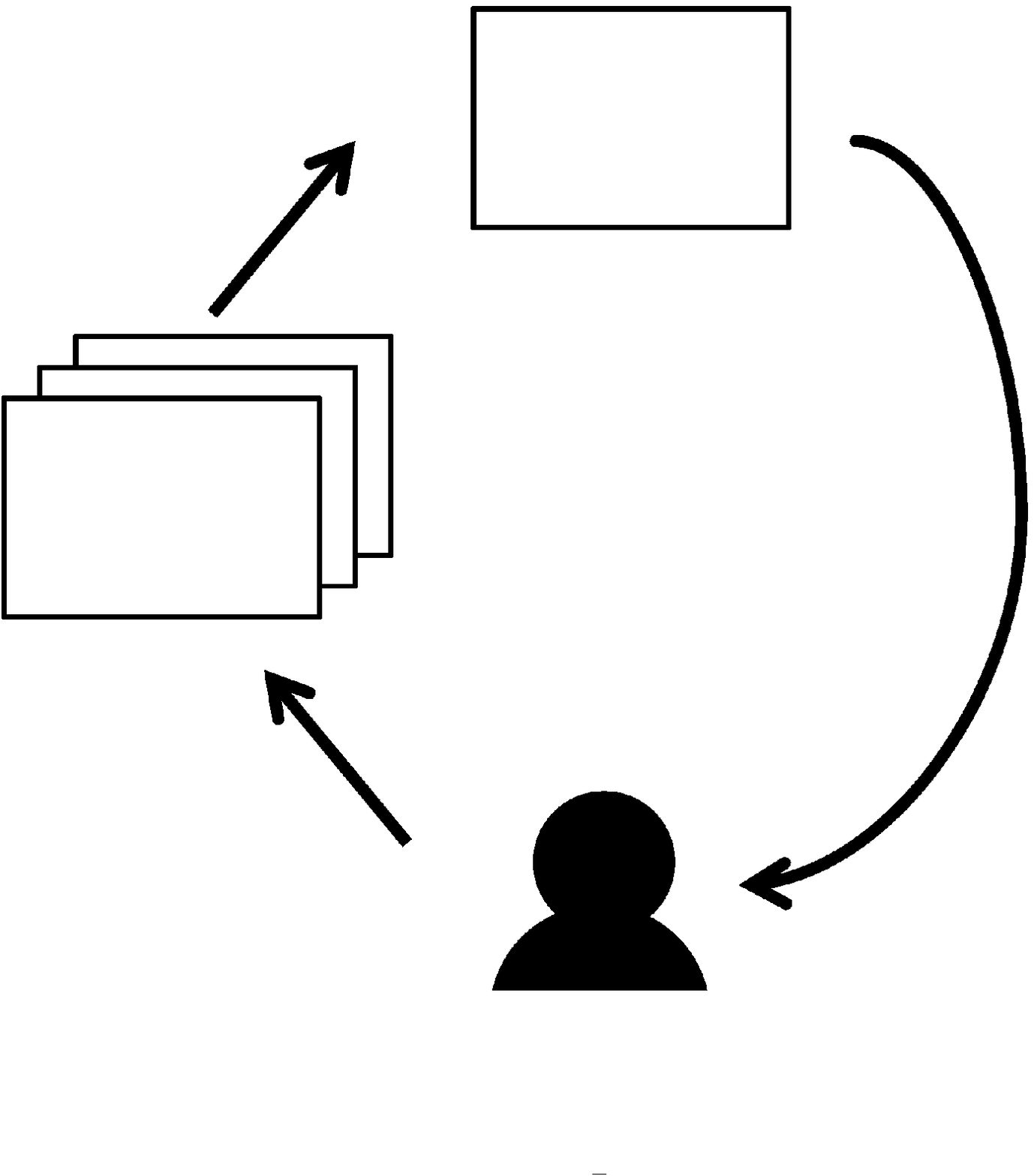}
		\put(-45,83){{\footnotesize Learner}}
		\put(-102,65){{\footnotesize Training}}
		\put(-5,20){{\footnotesize $x,?$}}
		\put(-65,20){{\footnotesize $x,y$}}
		\put(2,45){{\footnotesize GAN}}
	}
	\caption{(a) Pool-based active learning scenario. The learner selects samples for querying from a given unlabeled pool. (b) GAAL algorithm. The learner synthesizes samples for querying using GAN.}
	\label{fig:loop}
\end{figure}

In the classification setting, given a pool of unlabeled data samples and a fixed labeling budget, active learning algorithms typically choose training samples strategically from a pool to maximize the accuracy of trained classifiers.
The goal of these algorithms is to reduce label complexity.
Such approaches are called pool-based active learning. This pool-based active learning approach is illustrated in Figure~\ref{fig:loop} (a).

In a nutshell, we propose to use GANs to synthesize informative training instances that are adapted to the current learner. We then ask human oracles to label these instances. The labeled data is added back to the training set to update the learner. This protocol is executed iteratively until the label budget is reached. This process is shown in Figure~\ref{fig:loop} (b).

The main contributions of this work are as follows:
\begin{itemize}
	\item To the best of our knowledge, this is the first active learning framework using deep generative models\footnote{The appendix of \cite{papernot2016semi} mentioned three active learning attempts but did not report numerical results. Our approach is also different from those attempts.}.
	\item While we do not claim our method is always superior to the previous active learners in terms of accuracy, in some cases, it yields classification performance not achievable even by a fully supervised learning scheme. With enough capacity from the trained generator, our method allows us to have control over the generated instances which may not be available to the previous active learners.
	\item We conduct experiments to compare our active learning approach with self-taught learning\footnote{See the supplementary document.}. The results are promising.
	\item  This is the first work to report numerical results in active learning synthesis for image classification. See \cite{Settles2010,Lang1992}. The proposed framework may inspire future GAN applications in active learning.
	\item The proposed approach should not be understood as a pool-based active learning method. Instead, it is active learning by query synthesis. We show that our approach can perform competitively when compared against pool-based methods.
\end{itemize}

\section{Related Work}
% How is this work related to previous works? put in historical context
\label{ch:related}
Our work is related to two different subjects, active learning and deep generative models.

Active learning algorithms can be categorized into stream-based, pool-based and learning by query synthesis.
Historically, stream-based and pool-based are the two popular scenarios of active learning \cite{Settles2010}. 

\begin{figure}[ht]
	%\vskip -0.1in
	\begin{center}
		\centerline{\includegraphics[width=0.24\columnwidth]{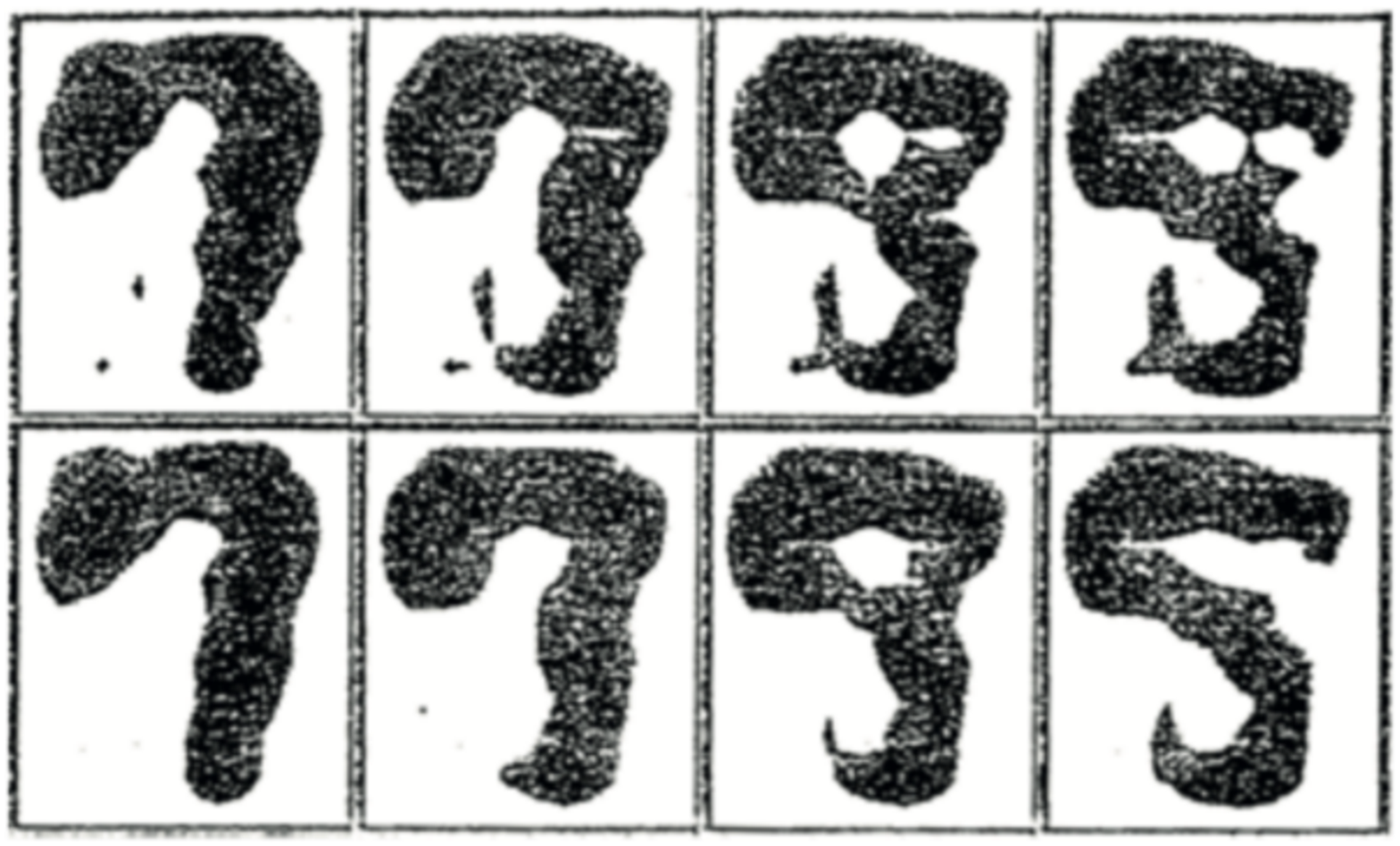}\;\;\;\;\;\;\;\;\;\;\;\;\;
		\includegraphics[width=0.28\columnwidth]{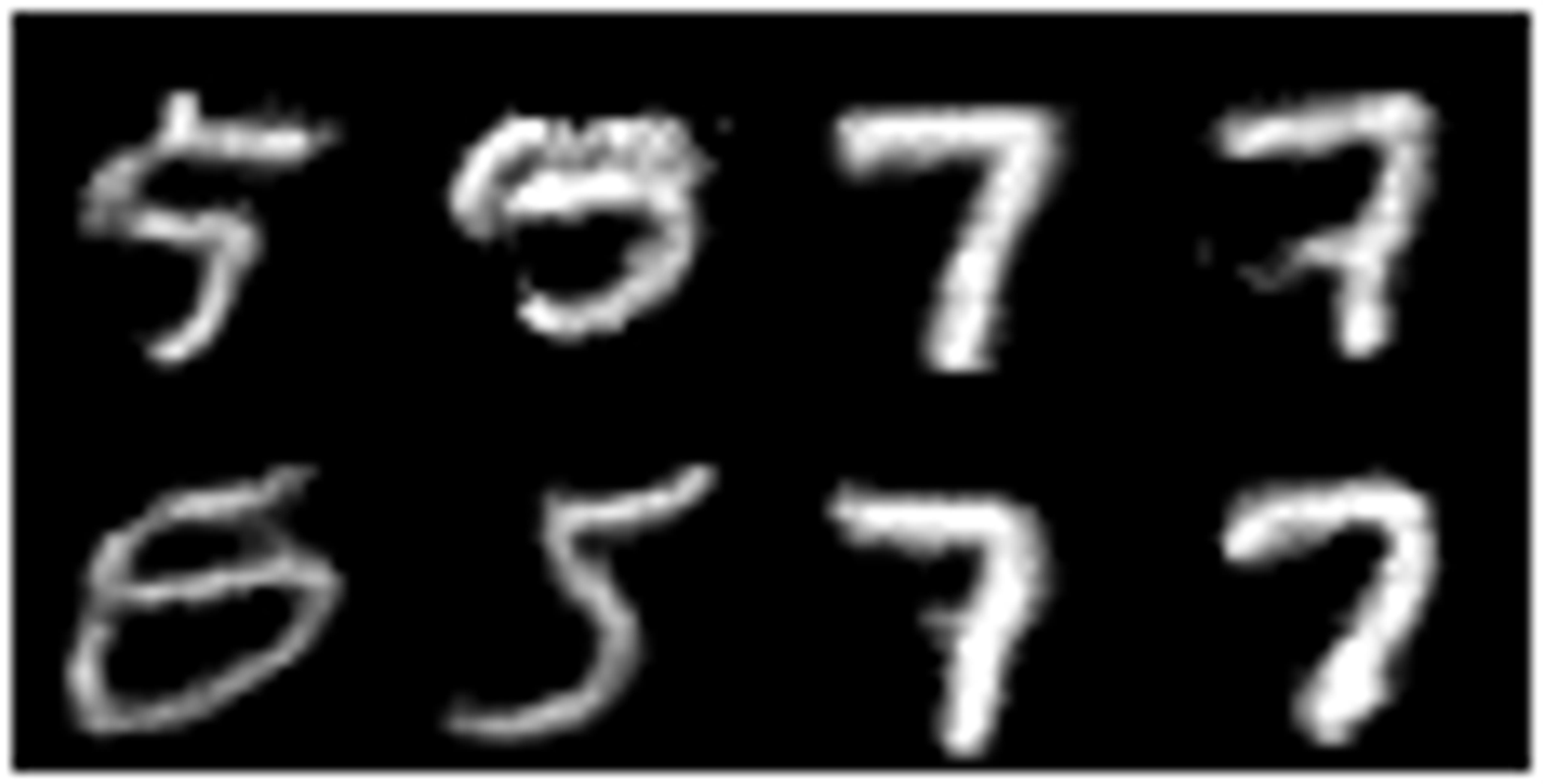}
		}
		\caption{(Left) Image queries synthesized by a neural network for handwritten digits recognition. Source: \cite{Lang1992}. (Right) Image queries synthesized by our algorithm, GAAL.}
		\label{fig:compare_lang}
	\end{center}
	\vskip -0.35in
\end{figure}

Our method falls into the category of query synthesis. Early active learning by queries synthesis achieves good results only in simple domains such as $X=\{0,1\}^3$, see \cite{Angluin1988,Angluin2001}. 
In \cite{Lang1992}, the authors synthesized learning queries and used human oracles to train a neural network for classifying handwritten characters. However, they reported poor results due to the images generated by the learner being sometimes unrecognizable to the human oracles. We will report results on similar tasks such as differentiating 5 versus 7, showing the advancement of our active learning scheme. Figure~\ref{fig:compare_lang} compares image samples generated by the method in \cite{Lang1992} and our algorithm.

The popular SVM$_{active}$ algorithm from \cite{Tong1998} is an efficient pool-based active learning scheme for SVM. Their scheme is a special instance of the uncertainty sampling principle which we also employ.
\cite{Jain2010} reduces the exhaustive scanning through database employed by SVM$_{active}$. Our algorithm shares the same advantage of not needing to test every sample in the database at each iteration of active learning. Although we do so by not using a pool at all instead of a clever trick.
\cite{wang2014active} proposed active transfer learning which is reminiscent to our experiments in Section~\ref{sec:al}. However, we do not consider collecting new labeled data in target domains of transfer learning. 

There have been some applications of generative models in semi-supervised learning and active learning. Previously, \cite{Nigam2000} proposed a semi-supervised learning approach to text classification based on generative models. \cite{Hospedales2013} applied Gaussian mixture models to active learning. In that work, the generative model served as a classifier.
Compared with these approaches, we apply generative models to directly synthesize training data. This is a more challenging task.
%To the best of our knowledge, this is the first active learning work using deep generative models.

One building block of our algorithm is the groundbreaking work of the GAN model in \cite{goodfellow2014generative}. Our approach is an application of GAN in active learning.

Our approach is also related to \cite{Springenberg2015} which studied GAN in a semi-supervised setting. However, our task is active learning which is different from the semi-supervised learning they discussed.
% previous active learning GAN
Our work shares the common strength with the self-taught learning algorithm in \cite{Raina2007} as both methods use the unlabeled data to help with the task. In the supplementary document, we compare our algorithm with a self-taught learning algorithm.

%%%
% adversarial training
In a way, the proposed approach can be viewed as an adversarial training procedure \cite{goodfellow2014explaining}, where the classifier is iteratively trained on the adversarial example generated by the algorithm based on solving an optimization problem.
\cite{goodfellow2014explaining} focuses on the adversarial examples that are generated by perturbing the original datasets within the small epsilon-ball whereas we seek to produce examples using active learning criterion.
%%%

To the best of our knowledge, the only previous mentioning of using GAN for active learning is in the appendix of \cite{papernot2016semi}. The authors discussed therein three attempts to reduce the number of queries. In the third attempt, they generated synthetic samples and sorted them by the information content whereas we adaptively generate new queries by solving an optimization problem. There were no reported active learning numerical results in that work.
\section{Background}
\label{background}
We briefly introduce some important concepts in active learning and generative adversarial network.
\subsection{Active Learning}
\label{bg:al}
In the PAC learning framework \cite{Valiant1984}, label complexity describes the number of labeled instances needed to find a hypothesis with error $\epsilon$. The label complexity of passive supervised learning, i.e. using all the labeled samples as training data, is $\mathcal{O}(\frac{d}{\epsilon})$ \cite{Vapnik1998}, where $d$ is the VC dimension of the hypothesis class $\mathcal{H}$.
Active learning aims to reduce the label complexity by choosing the most informative instances for querying while attaining low error rate.
For example, \cite{Hanneke2007} proved that the active learning algorithm from \cite{Cohn1994} has the label complexity bound $\mathcal{O}(\theta d \log\frac{1}{\epsilon})$, where $\theta$ is defined therein as the disagreement coefficient, thus reducing the theoretical bound for the number of labeled instances needed from passive supervised learning.
Theoretically speaking, the asymptotic accuracy of an active learning algorithm can not exceed that of a supervised learning algorithm. In practice, as we will demonstrate in the experiments, our algorithm may be able to achieve higher accuracy than the passive supervised learning in some cases.

Stream-based active learning makes decisions on whether to query the streamed-in instances or not. Typical methods include \cite{Beygelzimer2008,Cohn1994,Dasgupta2007}. In this work, we will focus on comparing pool-based and query synthesis methods.

In pool-based active learning, the learner selects the unlabeled instances from an existing pool based on a certain criterion. 
Some pool-based algorithms make selections by using clustering techniques or maximizing a diversity measure, e.g. \cite{Brinker,Xu2007,Dasgupta2008,Nguyen,Yang2015,Hoi2009}. 
Another commonly used pool-based active learning principle is uncertainty sampling. It amounts to querying the most uncertain instances. For example, algorithms in \cite{Tong1998,Campbell2000} query the labels of the instances that are closest to the decision boundary of the support vector machine. 
Figure~\ref{fig:gaal} (a) illustrates this selection process.
Other pool-based works include \cite{houlsby2012collaborative} which proposes a Bayesian active learning by disagreement algorithm in the context of learning user preferences, \cite{guillory2010interactive,golovin2010adaptive} which study the submodularity nature of sequential active learning schemes.

Mathematically, let $P$ be the pool of unlabeled instances, and $f=W \phi(x)+b$ be the separating hyperplane. $\phi$ is the feature map induced by the SVM kernel. The SVM$_{active}$ algorithm in \cite{Tong1998} chooses a new instance to query by minimizing the distance (or its proxy) to the hyperplane
\begin{equation}
\label{eqn:svmactive}
\min_{x\in P} \|W\phi(x)+ b\|.
\end{equation}
This formulation can be justified by the version space theory in separable cases \cite{Tong1998} or by other analyses in non-separable cases, e.g., \cite{Campbell2000,Bordes2005}.
This simple and effective method is widely applied in many studies, e.g., \cite{Goh2004,Warmuth2002}.

In the query synthesis scenario, an instance $x$ is synthesized instead of being selected from an existing pool.  Previous methods tend to work in simple low-dimensional domains \cite{Angluin2001} but fail in more complicated domains such as images \cite{Lang1992}. Our approach aims to tackle this challenge.

For an introduction to active learning, readers are referred to \cite{Settles2010,Dasgupta2011}.
\subsection{Generative Adversarial Networks}
\label{bg:gan}
Generative adversarial networks (GAN) is a novel generative model invented by \cite{goodfellow2014generative}.
It can be viewed as the following two-player minimax game between the generator $G$ and the discriminator $D$,
\begin{equation}
\label{eqn:gangame}
\min_{\theta_2}\max_{\theta_1} \Big\{{\mathbb{E}}_{x\sim p_\text{data}}\log D_{\theta_1}(x) + {\mathbb{E}}_{z}\log (1-D_{\theta_1}(G_{\theta_2}(z)))\Big\},
\end{equation}
where $p_\text{data}$ is the underlying distribution of the real data and $z$ is uniformly distributed random variable. $D$ and $G$ each has its own set of parameter $\theta_1$ and $\theta_2$.
By solving this game, a generator $G$ is obtained. In the ideal scenario, given random input $z$, we have $G(z)\sim  p_\text{data}$.
However, finding this Nash Equilibrium is a difficult problem in practice. There is no theoretical guarantee for finding the Nash Equilibrium due to the non-convexity of $D$ and $G$. A gradient descent type algorithm is typically used for solving this optimization problem.

A few variants of GAN have been proposed since \cite{goodfellow2014generative}.
The authors of \cite{Radford2015} use GAN with deep convolutional neural network structures for applications in computer vision(DCGAN). DCGAN yields good results and is relatively stable.
Conditional GAN\cite{Gauthier2014,Dosovitskiy2014,Mirza2014} is another variant of GAN in which the generator and discriminator can be conditioned on other variables, e.g., the labels of images. Such generators can be controlled to generate samples from a certain category.
\cite{Chen2016} proposed infoGAN which learns disentangled representations using unsupervised learning.

A few updated GAN models have been proposed. \cite{Salimans2016} proposed a few improved techniques for training GAN. Another potentially important improvement of GAN, Wasserstein GAN, has been proposed by \cite{Arjovsky2017,gulrajani2017improved}. The authors proposed an alternative to training GAN which can avoid instabilities such as mode collapse with theoretical analysis. They also proposed a metric to evaluate the quality of the generation which may be useful for future GAN studies. Possible applications of Wasserstein GAN to our active learning framework are left for future work.

The invention of GAN triggered various novel applications.
\cite{Yeh2016} performed image inpainting task using GAN.
\cite{Zhu2016} proposed iGAN to turn sketches into realistic images.
\cite{Ledig2016} applied GAN to single image super-resolution.
\cite{zhu2017unpaired} proposed CycleGAN for image-to-image translation using only unpaired training data.

Our study is the first GAN application to active learning.

For a comprehensive review of GAN, readers are referred to \cite{Goodfellow-et-al-2016}.
\section{Generative Adversarial Active Learning}
\label{sec:gaal}
In this section, we introduce our active learning approach which we call Generative Adversarial Active Learning (GAAL). It combines query synthesis with the uncertainty sampling principle.

\begin{figure}[ht]
	\centering
	\subfigure[SVM$_{active}$]{%
		\includegraphics[width=0.14\columnwidth]{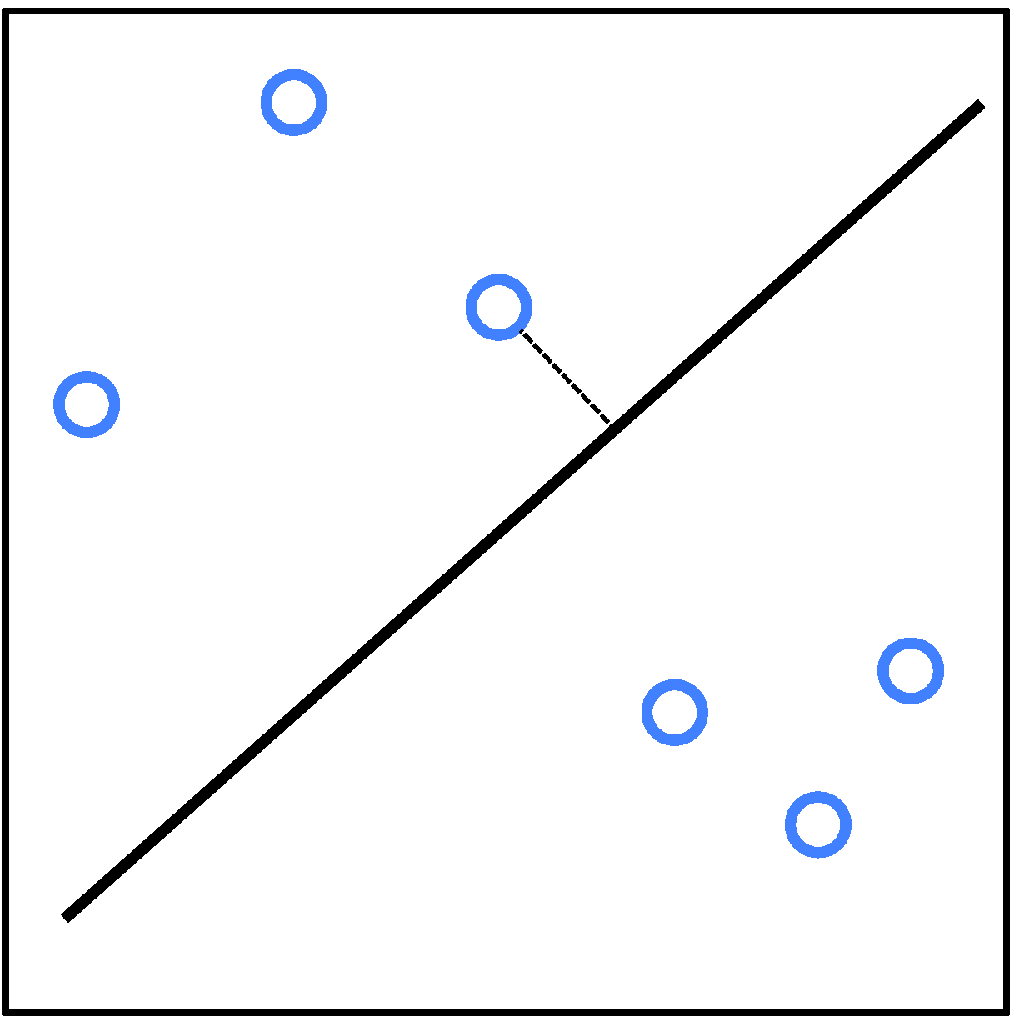}
	}\;\;\;\;\;\;\;\;\;\;\;\;\;\;\;\;\;\;\;\;\;\;\;\;\;\;\;\;\;\;
	\subfigure[GAAL]{%
		\includegraphics[width=0.14\columnwidth]{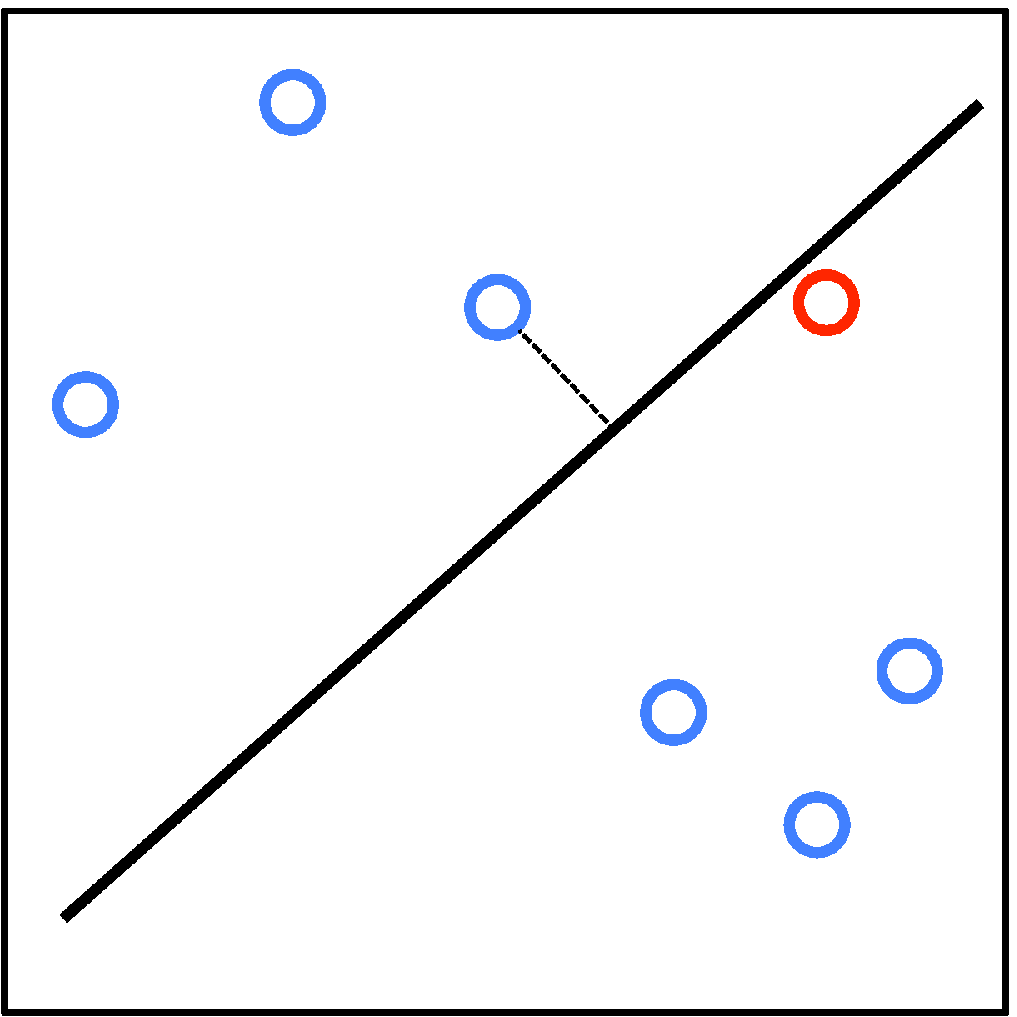}
	}
	\caption{(a) SVM$_{active}$ algorithm selects the instances that are closest to the boundary to query the oracle. (b) GAAL algorithm synthesizes instances that are informative to the current learner. Synthesized instances may be more informative to the learner than other instances in the existing pool.}
	\label{fig:gaal}
\end{figure}

The intuition of our approach is to generate instances which the current learner is uncertain about, i.e. applying the uncertainty sampling principle. 
%To this end, we formulate the optimization problem
%\begin{equation}
%\label{eqn:generalloss}
%\min _z \Big\{\mathrm{L}_\mathrm{active} (G(z)) + \mathrm{L} _\mathrm{reg} (G(z))\Big\},
%\end{equation}
%where $z$ is the latent variable and $G$ is obtained by the GAN algorithm.
%The first term $\mathrm{L}_\mathrm{active} (G(z))$ is the loss function for generating an informative active learning query. A small value of $\mathrm{L}_\mathrm{active} (G(z))$ indicates the generated instance $G(z)$ is informative to the learner.
%The second term $\mathrm{L} _\mathrm{reg} (G(z))$ is a regularization term which ensures the quality of generated samples. In the aforementioned adversarial setting, it penalizes low-quality samples.
One particular choice for the loss function is based on uncertainty sampling principle explained in section \ref{bg:al}.
%This specific adaptation of uncertainty sampling in this work may be better coined as uncertainty generation to indicate it is not a pool-based sampling scheme.
In the setting of a classifier with the decision function $f(x) = W\phi(x) + b$, the (proxy) distance to the decision boundary is $\|W\phi(x) + b\|$.
Similar to the intuition of \eqref{eqn:svmactive}, given a trained generator function $G$, we formulate the active learning synthesis as the following optimization problem
\begin{equation}
\label{eqn:gaal}
\min_z \|W^\top \phi(G(z)) + b\|,
\end{equation}
where $z$ is the latent variable and $G$ is obtained by the GAN algorithm. Intuitively, minimizing this loss will push the generated samples toward the decision boundary. Figure~\ref{fig:gaal} (b) illustrates this idea. Compared with the pool-base active learning in Figure~\ref{fig:gaal} (a), our hope is that it may be able to generate more informative instances than those available in the existing pool.

The solution(s) to this optimization problem, $G(z)$, after being labeled, will be used as new training data for the next iteration. We outline our procedure in Algorithm~\ref{alg:loop}.
\begin{algorithm}[tb]
	\caption{Generative Adversarial Active Learning (GAAL)}
	\label{alg:loop}
	\begin{algorithmic}[1]
		\STATE Train generator $G$ on all unlabeled data by solving \eqref{eqn:gangame}
		\STATE Initialize labeled training dataset $\mathcal{S}$ by randomly picking a small fraction of the data to label
		\REPEAT
		\STATE Solve optimization problem \eqref{eqn:gaal} according to the current learner by descending the gradient $$\nabla_z\|W^\top \phi(G(z)) + b\|$$
		\STATE Use the solution $\{z_1, z_2,\dots\}$ and $G$ to generate instances for querying
		\STATE Label $\{G(z_1), G(z_2),\dots\}$ by human oracles
		\STATE Add labeled data to the training dataset $\mathcal{S}$ and re-train the learner, update $W$, $b$
		\UNTIL{Labeling budget is reached}
	\end{algorithmic}
\end{algorithm}
%
%Formulation \eqref{eqn:generalloss} offers the flexibility of using other loss terms.
%For examples, in the case of logistic regression as the classifier of choice, uncertain sampling principle in active learning corresponds to the active loss term choice
%%\begin{align}
%%\label{eqn:logloss}
%%\begin{split}
%$\mathrm{L}_\mathrm{active} (z)= - h\left(G\left(z\right)\right) \log h\left(G\left(z\right)\right)
%-\left(1-h\left(G\left(z\right)\right)\right)\log \left(1-h\left(G\left(z\right)\right)\right)$,
%%\end{split}
%%\end{align}
%where $h(x)=\frac{1}{1+e^{-\theta x}}$ is the logistic function. $\theta$ is the parameter of the classifier, similar to $W, b$ in \eqref{eqn:gaal}. The derivation of this formulation is analogous to the entropy measure in \cite{Joshi2009}.
It is possible to use a state-of-the-art classifier, such as convolutional neural networks. To do this, we can replace the feature map $\phi$ in Equation~\ref{eqn:gaal} with a feed-forward function of a convolutional neural network. In that case, the linear SVM will become the output layer of the network.
In step 4 of Algorithm~\ref{alg:loop}, one may also use a different active learning criterion. We emphasis that our contribution is the general framework instead of a specific criterion.

In training GAN, we follow the procedure detailed in \cite{Radford2015}.
Optimization problem \eqref{eqn:gaal} is non-convex with possibly many local minima. One typically aims at finding good local minima rather than the global minimum. We use a gradient descent algorithm with momentum to solve this problem. We also periodically restart the gradient descent to find other solutions. The gradient of $D$ and $G$ is calculated using back-propagation.

Alternatively, we can incorporate diversity into our
active learning principle. Some active learning approaches rely on maximizing diversity measures, such as the Shannon Entropy. In our case, we can include in the objective function \eqref{eqn:gaal} a diversity measure such as proposed in \cite{Yang2015,Hoi2009}, thus increasing the diversity of samples. The evaluation of this alternative approach is left for future work.
%\begin{equation}
%\label{eqn:entropygaall}
%\min_z \{\mathbb{E}[\mathcal{H}(G(z)) ] + \lambda\log (1-D(G(z)))\},
%\end{equation}
%where $\mathcal{H}(G(z))$ is the Shannon Entropy of a certain instance $G(z)$. Minimizing the first term corresponds to maximizing the Shannon Entropy of the generated samples. How the expectation and probability are estimated in practice is left for future work. This work focus on the the application of uncertainty sampling.
\section{Experiments}
\label{sec:exp}
We perform active learning experiments using the proposed approach. 
We also compare our approach to self-taught learning, a type of transfer learning method, in the supplementary document.
The GAN implementation used in our experiment is a modification of a publicly available TensoFlow DCGAN implementation\footnote{https://github.com/carpedm20/DCGAN-tensorflow}. 
The network architecture of DCGAN is described in \cite{Radford2015}.

In our experiments, we focus on binary image classification. Although this can be generalized to multiple classes using one-vs-one or one-vs-all scheme \cite{Joshi2009}. Recent advancements in GAN study show it could potentially model language as well \cite{gulrajani2017improved}. Although those results are preliminary at the current stage. We use a linear SVM as our classifier of choice (with parameter $\gamma=0.001$). Even though classifiers with much higher accuracy (e.g., convolutional neural networks) can be used, our purpose is not to achieve absolute high accuracy but to study the relative performance between different active learning schemes.

The following schemes are implemented and compared in our experiments.
\vspace{-0.2cm}
\begin{itemize}
	\item The proposed generative adversarial active learning (GAAL) algorithm as in Algorithm~\ref{alg:loop}.
	\item Using regular GAN to generate training data. We refer to this as simple GAN.
	\item SVM$_{active}$ algorithm from \cite{Tong1998}.
	\item Passive random sampling, which randomly samples instances from the unlabeled pool.
	\item Passive supervised learning, i.e., using all the samples in the pool to train the classifier.
	\item Self-taught learning from \cite{Raina2007}.
\end{itemize}
\vspace{-0.2cm}
We initialize the training set with 50 randomly selected samples.
The algorithms proceed with a batch of 10 queries every time.

\begin{figure}[ht]
	%\vskip -0.1in
	\begin{center}
		\centerline{\includegraphics[width=0.28\columnwidth]{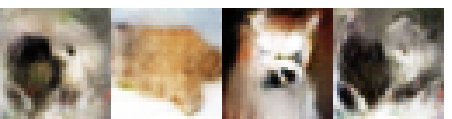}}
		\vspace{0.1cm}
		\centerline{\includegraphics[width=0.31\columnwidth]{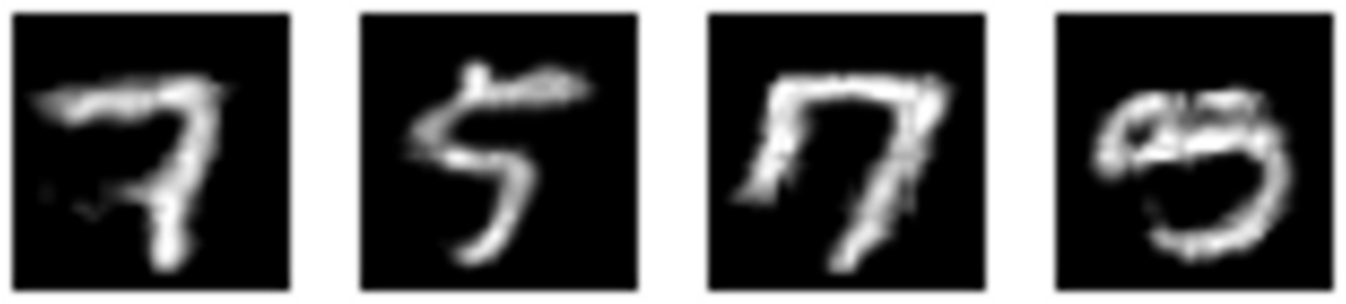}\;\;\;
		\includegraphics[width=0.31\columnwidth]{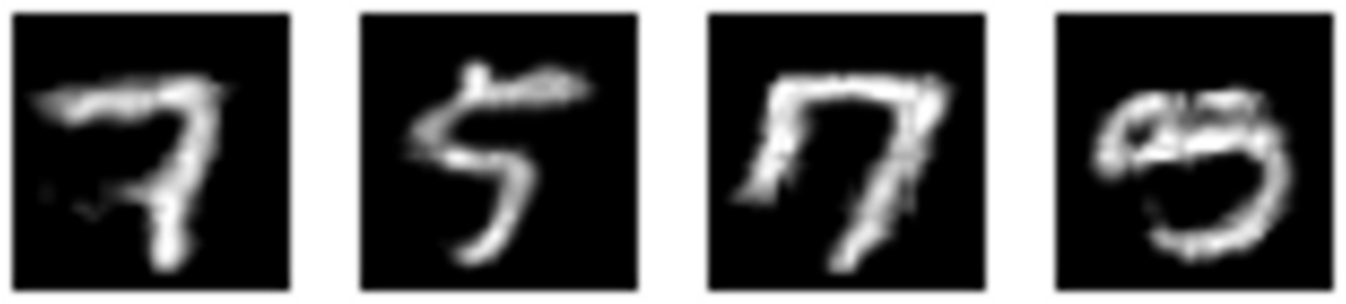}
		}
		\caption{Samples generated by GAAL (Top) Generated samples in cat and dog categories. (Bottom Left) MNIST dataset. (Bottom Right)  CIFAR-10 dataset.}
		\label{fig:gen_samples}
	\end{center}
	\vskip -0.45in
\end{figure}

We use two datasets for training, the MNIST and CIFAR-10. The MNIST dataset is a well-known image classification dataset with 60000 training samples. The training set and the test set follow the same distribution. We perform the binary classification experiment distinguishing 5 and 7 which is reminiscent to \cite{Lang1992}.
The training set of CIFAR-10 dataset consists of 50000 $32\times32$ color images from 10 categories. One might speculate the possibility of distinguishing cats and dogs by training on cat-like dogs or dog-like cats. In practice, our human labelers failed to confidently identify most of the generated cat and dog images. Figure~\ref{fig:gen_samples} (Top) shows generated samples. The authors of \cite{Salimans2016} reported attempts to generate high-resolution animal pictures, but with the wrong anatomy.
We leave this task for future studies, possibly with improved techniques such as \cite{Arjovsky2017,gulrajani2017improved}.
%\begin{figure}[ht]
%	%	\vskip -0.2in
%	\begin{center}
%		\centerline{\includegraphics[width=0.28\columnwidth]{fig/cat_dog}}
%		\caption{Generated samples in cat and dog categories}
%		\label{fig:cat_dog}
%	\end{center}
%	\vskip -0.2in
%\end{figure} 
For this reason, we perform binary classification on the automobile and horse categories. It is relatively easy for human labelers to identity car and horse body shapes.
Typical generated samples, which are presented to the human labelers, are shown in Figure~\ref{fig:gen_samples}.

\subsection{Active Learning}
\label{sec:al}
We use all the images of 5 and 7 from the MNIST training set as our unlabeled pool to train the generator $G$. Different from traditional active learning, we do not select new samples from the pool after initialization. Instead, we apply Algorithm~\ref{alg:loop} to generate a training query. For the generator $D$ and $G$, we follow the same network architecture of \cite{Radford2015}. We use linear SVM as our classifier although other classifiers can be used, e.g. \cite{Tong1998,Schein2007,Settles2010}. 

\begin{figure}[ht]
%	\vspace{-1.2cm}
	\begin{center}
		\centerline{\includegraphics[width=0.4\columnwidth]{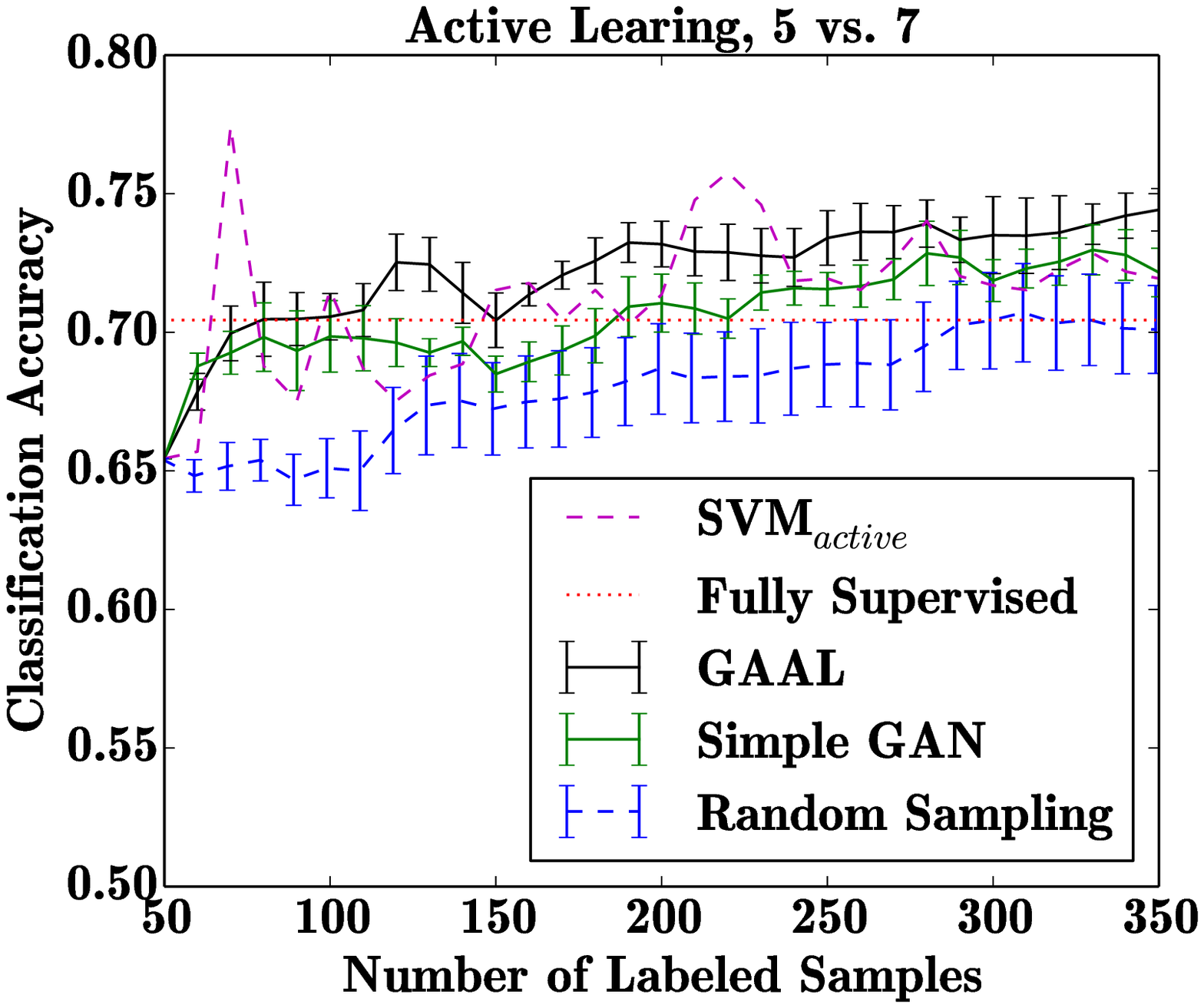}}
			
		\centerline{
			\includegraphics[width=0.35\columnwidth]{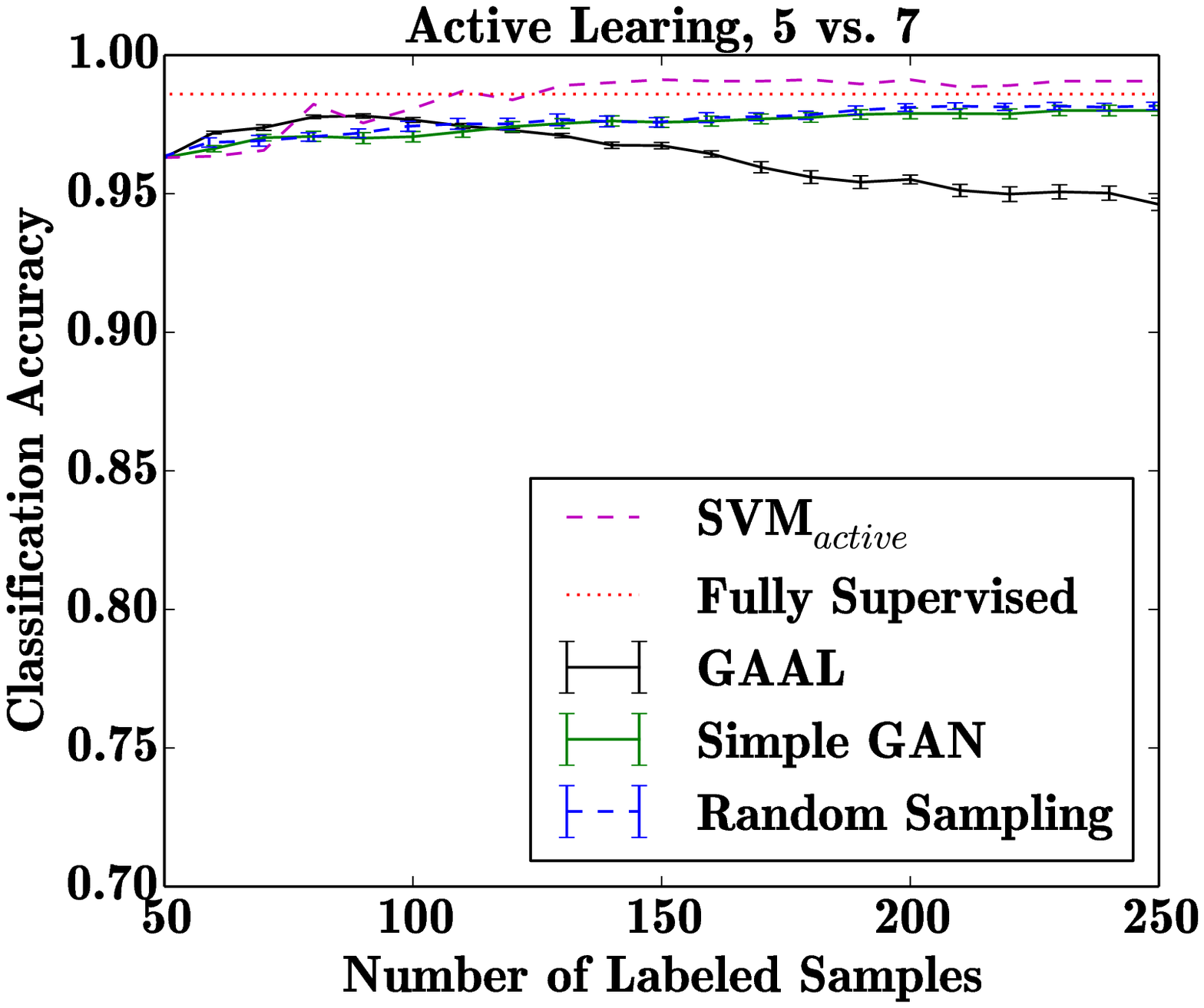}\;\;\;\;\;\;\;\;\;\;\;\;
			\includegraphics[width=0.35\columnwidth]{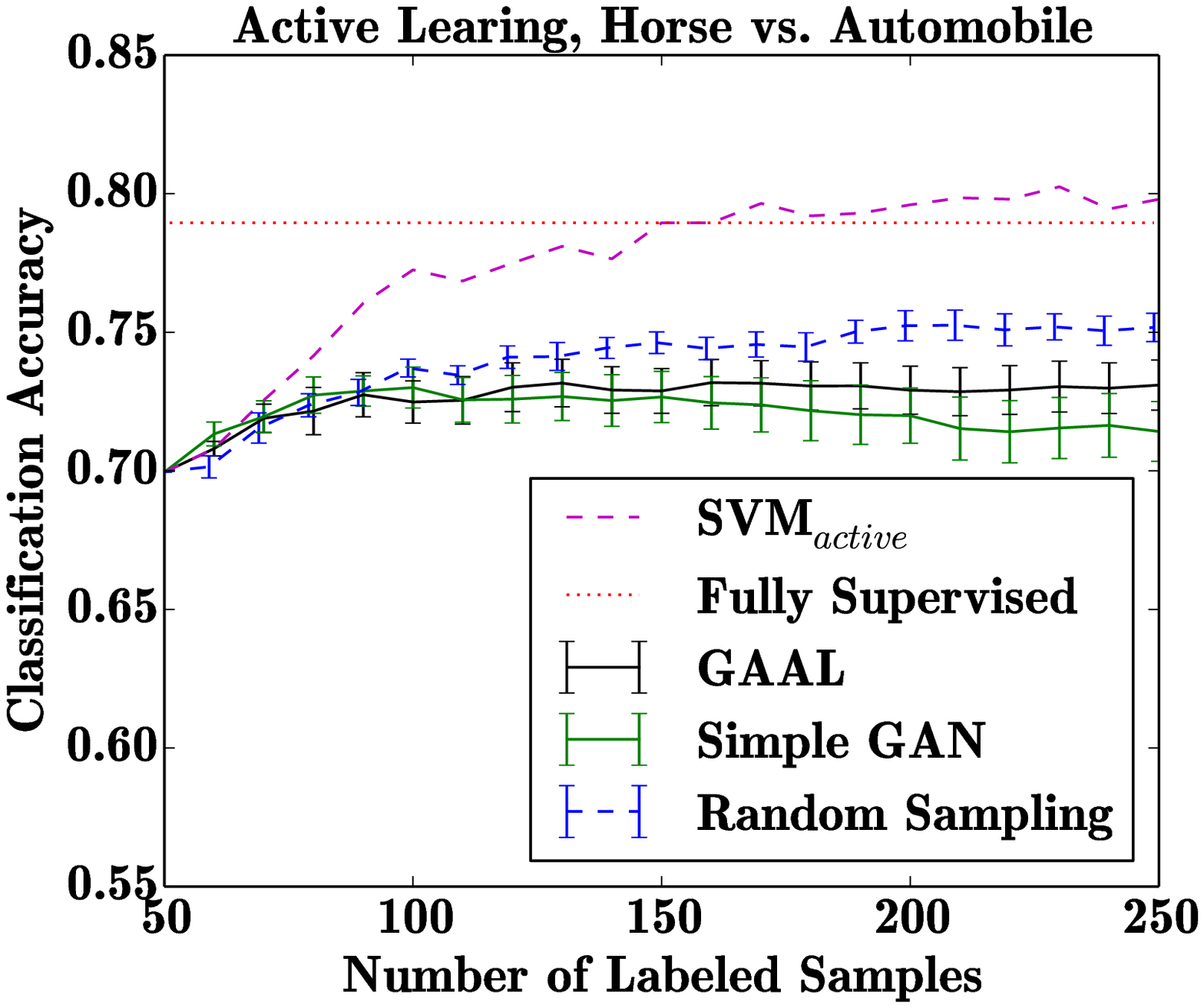}
		}
		\caption{Active learning results. (Top) Train on MNIST, test on USPS. Classifying 5 and 7. 
			The results are averaged over 10 runs. (Bottom Left) Train on MNIST, test on MNIST. Classifying 5 and 7. (Bottom Right) CIFAR-10 dataset, classifying automobile and horse. The results are averaged over 10 runs. The error bars represent the empirical standard deviation of the average values. The figures are best viewed in color.}
		\label{fig:alsucks}
	\end{center}
	\vspace{-0.5in}
\end{figure}

We first test the trained classifier on a test set that follows a  distribution different from the training set. 
One purpose is to demonstrate the adaptive capability of the GAAL algorithm. 
In addition, because the MNIST test set and training set follow the same distribution, pool-based active learning methods have an natural advantage over active learning by synthesis since they use real images drawn from the exact same distribution as the test set. 
It is thus reasonable to test on sets that follow different, albeit similar, distributions.
To this end, we use the USPS dataset from \cite{LeCun1989} as the test set with standard preprocessing. 
In reality, such settings are very common, e.g., training autonomous drivers on simulated datasets and testing on real vehicles; training on handwriting characters and recognizing writings in different styles, etc.
This test setting is related to transfer learning, where the distribution of the training domain $P_{tr}(x,y)$ is different from that of the target domain $P_{te}(x,y)$.
%It is also related to the self-taught learning setting which we discuss in a later experiment.
Figure~\ref{fig:alsucks} (Top) shows the results of our first experiment.
%\begin{figure}[ht]
%	\vspace{-0.5cm}
%	\begin{center}\centerline{\includegraphics[width=0.4\columnwidth]{fig/redo_mnist}
%	}
%		\caption{Active learning results. Train on MNIST, test on USPS. Classifying 5 and 7. 
%		The results are averaged over 10 runs. The error bars represent the empirical standard deviation of the average values. The figure is best viewed in color.}
%		\label{fig:transfer}
%	\end{center}
%\end{figure}

When using the full training set, with 11000 training images, the fully supervised accuracy is at $70.44\%$. The accuracy of the random sampling scheme steadily approaches that level.
On the other hand, GAAL is able to achieve accuracies better than that of the fully supervised scheme. With 350 training samples, its accuracy improves over supervised learning and even SVM$_{active}$, an aggressive active learner \cite{dasgupta2005analysis,Tong1998}. 
Obviously, the accuracy of both SVM$_{active}$ and random sampling will eventually converge to the fully supervised learning accuracy. Note that for the SVM$_{active}$ algorithm, an exhaustive scan through the training pool is not always practical. In such cases, the common practice is to restrict the selection pool to a small random subset of the original data.

For completeness, we also perform the experiments in the settings where the training and test set follow the same distribution. Figure~\ref{fig:alsucks} (Bottom) shows these results. 
%
%
%\begin{figure}[ht]
%\vspace{-0.3cm}
%	\centering
%	\subfigure{%
%		\includegraphics[width=0.4\columnwidth]{fig/redo_mnist_test}
%	}\;\;\;\;\;\;\;\;\;\;\;\;\;\;\;
%	\subfigure{%
%		\includegraphics[width=0.4\columnwidth]{fig/redo_cifar}
%	}
%	\caption{(Left) MNIST dataset, classifying 5 and 7 (Right) CIFAR-10 dataset, classifying automobile and horse. Results are averaged over 10 runs. The error bars represent the empirical standard deviation of the average values.}
%	\label{fig:alsucks}
%\end{figure}
%We notice that the performance of GAAL is not as good as the pool-based approach.
Somewhat surprisingly, in Figure~\ref{fig:alsucks} (Left), GAAL's classification accuracy starts to drop after about 100 samples.
One possible explanation is that GAAL may be generating 
points close to the boundary that are also close to each other. This is more likely to happen if the boundary does not change much from one active learning cycle to the next. This probably happens because the test and train sets are the identically distributed and simple, like MNIST.
Therefore, after a while, the training set may be filled with many similar points, biasing the classifier and hurting accuracy.
In contrast, because of the finite and discrete nature of pools in the given datasets, a pool-based approach, such as SVM$_{active}$, most likely explores points near the boundary that are substantially
different. It is also forced to explore further points once these close-by points have already been selected. In a sense, the strength of GAAL might in fact be hurting its classification accuracy. We believe this effect is not so pronounced when the test and train sets are different because the boundary changes more significantly from one cycle to the next, which in turn induces some diversity in the generated samples.

To reach competitive accuracy when the training and test set follow the same distribution, we might  incorporate a diversity term into our objective function in GAAL.
We will address this in future work.
In the CIFAR-10 dataset, our human labeler noticed higher chances of bad generated samples, e.g., instances fail to represent either of the categories. This may be because of the significantly higher dimensions than the MNIST dataset. In such cases, we asked the labelers to only label the samples they can distinguish. We speculate recent improvements on GAN, e.g., \cite{Salimans2016,Arjovsky2017,gulrajani2017improved}, may help mitigate this issue given the cause is the instability of GANs. Addressing this limitation will be left to future studies.
%\subsection{CIFAR-10}

% Figure~\ref{fig:cifar} shows the results.
%\begin{figure}[ht]
%	%\vskip 0.2in
%	\begin{center}
%		\centerline{\includegraphics[width=0.6\columnwidth]{fig/carhorse}}
%		\caption{Active learning results of the CIFAR-10 dataset, classifying automobile and horse.}
%		\label{fig:cifar}
%	\end{center}
%	\vskip -0.3in
%\end{figure} 
%In this experiment, GAAL performs on par with the random sampling scheme and better than the passive GAN scheme. However, it is not able to beat Tong\&Koller's active learning algorithm. This may be because that higher dimensions require more active learning iterations for GAAL to perform better.  

%%%
% new
\subsection{Balancing exploitation and exploration}
\label{sec:ee}
\begin{figure}[h!]
	\begin{center}\centerline{\includegraphics[width=0.4\columnwidth]{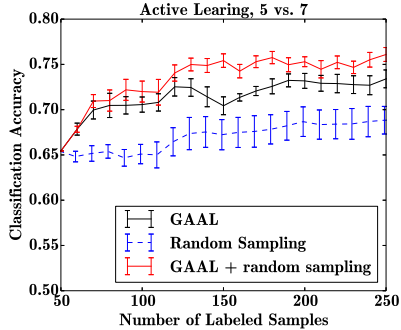}
		}
		\caption{Active learning results using a mixed scheme. The mixed scheme executes one iteration of random sampling after every five iterations of GAAL algorithm. Train on MNIST, test on USPS. Classifying 5 and 7. 
			The results are averaged over 10 runs. The error bars represent the empirical standard deviation of the average values. The figure is best viewed in color.}
		\label{fig:mixed}
	\end{center}
\end{figure}

The proposed Algorithm~\ref{alg:loop} can be understood as an exploitation method, i.e., it focuses on generating the most informative training data based on the current decision boundary On the other hand, it is often desirable for the algorithm to explore the new areas of the data. To achieve this, we modify Algorithm~\ref{alg:loop} by simply executing random sampling every once in a while. This is a common practice in active learning \cite{baram2004online,roder2012active}. We use the same experiment setup as in the previous section. Figure~\ref{fig:mixed} shows the results of this mixed scheme.

A mixed scheme is able to achieve better performance than either using GAAL or random sampling alone. Therefore, it implies that GAAL, as an exploitation scheme, performs even better in combination with an exploration scheme. A detailed analysis such mixed schemes will be an interesting future topic.
\section{Discussion and Future Work}
In this work, we proposed a new active learning approach, GAAL, that employs the generative adversarial networks.
One possible explanation for GAAL not outperforming the pool-based approaches in some settings is that, in traditional pool-based learning, the algorithm will eventually exhaust all the points near the decision boundary thus start exploring further points. However, this is the not the case in GAAL as it can always synthesize points near the boundary. This may in turn cause the generation of similar samples, thus reducing the effectiveness. We suspect incorporating a diversity measure into the GAAL framework as discussed at the end of Section~\ref{sec:gaal} might mitigate this issue. This issue is related to the exploitation and exploration trade-off which we explored in brief. 

The results of this work are enough to inspire future studies of deep generative models in active learning.
However, much work remains in establishing theoretical analysis and reaching better performance.
We also suspect that GAAL can be modified to generate adversarial examples such as in \cite{goodfellow2014explaining}.
The comparison of GAAL with transfer learning (see the supplementary document) is particularly interesting and worth further investigation.
We also plan to investigate the possibility of using Wasserstein GAN in our framework.

\bibliography{ref}
\bibliographystyle{plain}
\section*{Appendix: Comparison with Self-taught Learning}
	\label{sect:stl}
	One common strength of GAAL and self-taught learning \cite{Raina2007} is that both utilize the unlabeled data to help with the classification task.
	As we have seen in the MNIST experiment, our GAAL algorithm seems to be able to adapt to the learner. The results in this experiment are preliminary and not meant to be taken as comprehensive evaluations.
	
	In this case, the training domain is mostly unlabeled. Thus the method we compare with is self-taught learning \cite{Raina2007}. Similar to the algorithm in \cite{Le}, we use a Reconstruction Independent Component Analysis (RICA) model with a convolutional layer and a pooling layer. RICA is similar to a sparse autoencoder. Following standard self-taught learning procedures, We first train on the unlabeled pool dataset. Then we use trained RICA as the a feature extractor to obtain higher level features from randomly selected MNIST images. We then concatenate the features with the original image data to train the classifier. Finally, we test the trained classifier on the USPS dataset. We test the training size of $250$, $500$, $1000$, and $5000$. The reason of doing so is that deep learning type techniques are known to thrive in the abundance of training data. They may perform relatively poorly with limited amount of training data, as in the active learning scenarios. We run the experiments for 100 times and average the results. We use the same setting for the GAAL algorithm as in Section~\ref{sec:al}. The classifier we use is a linear SVM. Table~\ref{tb:selftaught} shows the classification accuracies of GAAL, self-taught learning and baseline supervised learning on raw image data.
	\begin{table}[h]
		\caption{Comparison of GAAL and self-taught learning}
		\label{tb:selftaught}
		\vskip 0.15in
		\begin{center}
			\begin{small}
				\begin{sc}
					\begin{tabular}{lcr}
						\hline
						%\abovestrut{0.20in}\belowspace
						Algoirthm & Training set size & accuracy \\
						\hline
						
						{\bf GAAL}    &  {\bf 250} & ${\bf76.42\%}$ \\
						Self-taught & 250 & $59.68\%$\\
						Supervised & 250 & $67.87\%$\\
						\hline
						Self-taught & 500 & $65.53\%$\\
						Supervised & 500 & $69.22\%$\\
						\hline
						Self-taught & 1000 & $71.96\%$\\
						Supervised & 1000 & $69.58\%$\\
						%					\hline\\
						%					Self-taught & 2000 & $75.84\%$\\
						%					Supervised & 2000 & $70.06\%$\\
						\hline
						{\bf SELF-TAUGHT} & {\bf 5000} & ${\bf 78.08\%}$\\
						Supervised & 5000 & $72.00\%$\\			
						\hline
					\end{tabular}
				\end{sc}
			\end{small}
		\end{center}
		\vskip -0.1in
	\end{table}
	Using GAAL on the raw features achieves a higher accuracy than that of the self-taught learning with the same training size of $250$. In fact, self-taught learning performs worse than the regular supervised learning when labeled data is scarce. This is possible for an autoencoder type algorithm. However, when we increase the training size, the self-taught learning starts to perform better. With 5000 training samples, self-taught learning outperforms GAAL with 250 training samples.
	
	Based on these results, we suspect that GAAL also has the potential to be used as a self-taught algorithm\footnote{At this stage, self-taught learning has the advantage that it can utilize any unlabeled training data, i.e., not necessarily from the categories of interest. GAAL does not have this feature yet.}. 
	In practice, the GAAL algorithm can also be applied on top of the features extracted by a self-taught algorithm. A comprehensive comparison with a more advanced self-taught learning method with deeper architecture is beyond the scope of this work.

%\newpage
%\begin{appendix}
%	\section{Hyperparameter tuning}
%	We conduct extensive experiment to examine the effect of different choices of $\lambda$ in Equation~\eqref{eqn:gaal}.
%	\begin{figure}[ht]
%		%\vskip 0.2in
%		\begin{center}
%			\centerline{\includegraphics[width=0.5\columnwidth]{fig/cnn_lambda}}
%			\caption{Accuracy plots of different hyperparameter $\lambda$ values. All results are averaged over 10 runs.}
%			\label{fig:cnn_lambda}
%		\end{center}
%		\vskip -0.3in
%	\end{figure} 
%	
%	\begin{figure}[ht]
%		%\vskip 0.2in
%		\begin{center}
%			\centerline{\includegraphics[width=0.5\columnwidth]{fig/cnn_auc}}
%			\caption{AUC plots of different hyperparameter $\lambda$ values. All results are averaged over 10 runs.}
%			\label{fig:cnn_auc}
%		\end{center}
%		\vskip -0.3in
%	\end{figure}
%\end{appendix}
\end{document}